\documentclass[sigconf,nonacm]{acmart}

\AtBeginDocument{%
  }

\usepackage{booktabs}
\usepackage{subcaption}
\usepackage{amsmath}
\begin{document}

\title{Incremental Recommendation via Causal Models}

\author{Athanasios Vlontzos}
\authornote{Work done while at Spotify.}
\affiliation{%
 \institution{Hologen}
  \institution{Imperial College London, UK}
  \country{}
}

\author{David Gustafsson}
\email{davidgustafsson@spotify.com}
\affiliation{%
  \institution{Spotify, Sweden}
  \country{}
}

\author{Michael O'Riordan}
\email{moriordan@spotify.com}
\affiliation{%
  \institution{Spotify, UK}
  \country{}
}

\author{Ciar\'an M. Gilligan-Lee}
\email{ciaranl@spotify.com}
\affiliation{%
  \institution{Spotify, Ireland}
  \country{}
}
\affiliation{%
  \institution{University College London, UK}
  \country{}
}

\begin{abstract}
Recommendation impressions are a finite resource, hence delivering a recommendation to a user who would discover the content organically yields no incremental value and displaces other recommendations that could. We address this by extending an existing production recommendation model to a causal architecture using holdback data that is already collected as part of routine experimentation infrastructure, requiring no new data collection.  A central challenge is that attribution windows differ between treated and holdback observations: treated users are attributed a stream within a short direct-response window, while holdback users are attributed organic streams over a multi-day window. This mismatch makes na\"ive treatment-effect subtraction invalid. We resolve this with a dual-threshold targeting policy that delivers a recommendation only when the probability of a treated stream is high \emph{and} the probability of organic stream is low.  In a production-scale A/B test on millions of Spotify users, this policy reduces recommendation impressions by 7\% with no statistically significant reduction in overall recommended content consumption. We further show that joint training with holdback data improves calibration of the treated head relative to the production baseline, and argue this can be taken as evidence that causal models learn more generalisable representations than models trained on observational data alone.
\end{abstract}




\maketitle

\section{Introduction}

Recommendation impressions are scarce. There is thus a strong incentive to ensure that each recommendation is \emph{efficient}. That is, that each recommendation has an impact. Standard recommenders are trained to maximize the probability a user streams the recommended content---a sensible objective that nevertheless ignores a critical question: \emph{would the user have streamed this content anyway?}

Users who would stream a piece of content regardless of receiving a recommendation are known in the causal literature as \emph{always-takers}~\cite{Rubin1974Estimating}. 
For a music streaming platform, an alwats-taker might be a user who closely follows an artist and checks their new releases weekly: delivering a recommendation for the artist's new album to such a user does not drive incremental consumption, it merely tags along with behavior that would occur organically. A model optimising stream probability will systematically favour these high-affinity users, because they are the easiest to predict, and will do so at the cost of ignoring users for whom a recommendation would make a genuine difference.

Correcting for this requires answering the counterfactual: what would the user have done \emph{without} the recommendation? This is the fundamental challenge of causal inference for recommendation. Many production recommendation systems already collect exactly the data needed to answer this question, in the form of holdback experiments, randomised trials in which a small fraction of eligible users are withheld recommendations. Holdback data provides direct samples from the counterfactual distribution of organic behavior, yet is rarely used to build causal recommendation models.

\begin{figure}[t]
  \includegraphics[width=\columnwidth]{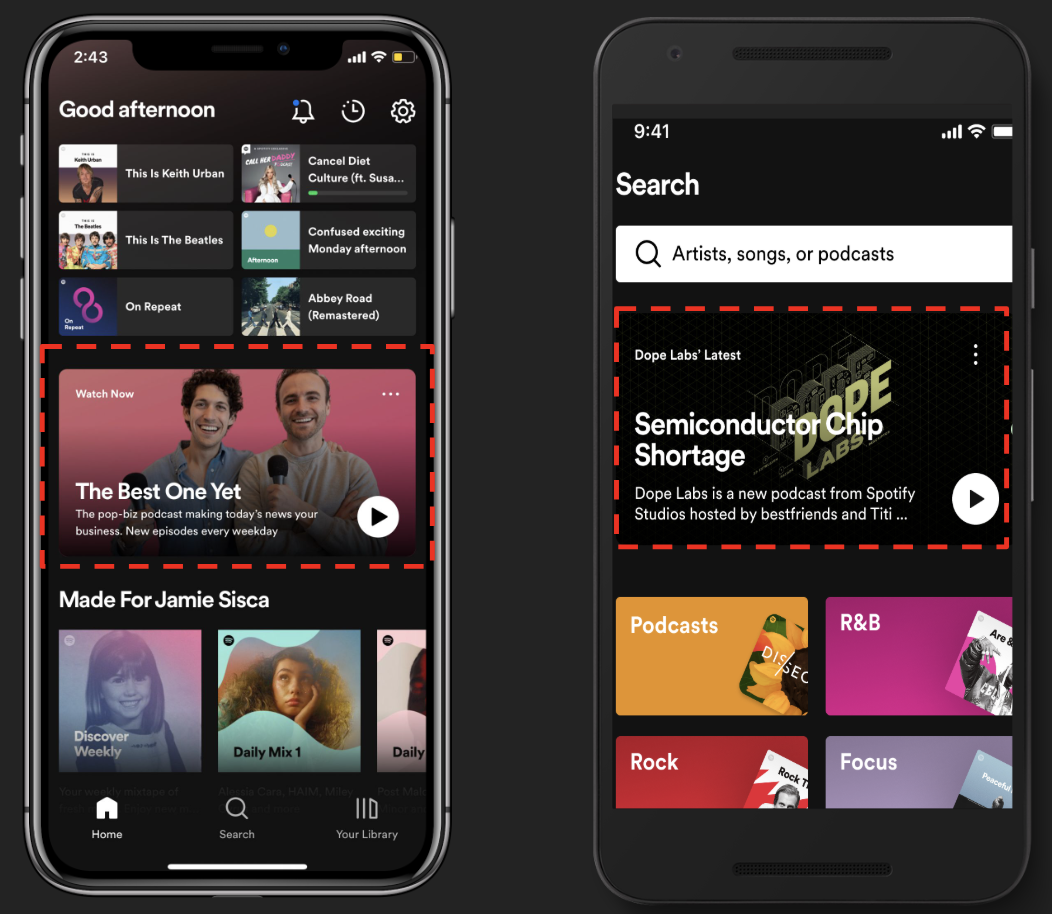}
  \caption{Recommendation surfaces in Spotify app where the model operates. Recommendation occupies slot indicated by the dashed red box.}
  \Description{Screenshot of the Spotify mobile app home screen showing recommendation banners and content cards in the feed.}
  \label{fig:ui}
\end{figure}

We show how to transform an existing production recommendation model into a causal incremental model using this holdback data, without any new data collection infrastructure. The key technical contribution is identifying and resolving a structural obstacle---the \emph{attribution window mismatch}---that prevents na\"ive treatment-effect estimation in this setting and, if ignored, would invalidate the resulting causal model. We validate the approach at production scale on Spotify's Home recommendation surface, illustrated in Figure~\ref{fig:ui}, demonstrating that the resulting dual-threshold targeting policy reduces recommendation impressions by 7\% with no statistically significant reduction in recommended content consumption.

Our contributions are as follows.
\begin{itemize}
  \item We identify and formalise the \textbf{attribution window mismatch}, a structural obstacle that prevents na\"ive conditional average treatment effect (CATE) estimation in recommendation systems with holdback groups, and propose a dual-threshold targeting policy as a principled resolution.
  \item We demonstrate how an existing production multi-task recommendation model can be extended to a causal Deep Twin Network architecture~\cite{Vlontzos2023TwinNets} using already-collected holdback data, requiring no new infrastructure.
  \item We validate this approach at production scale: a live A/B test on millions of Spotify users demonstrates a 7\% reduction in recommendation impressions with no significant impact on recommended content consumption.
  \item We show that joint training with holdback data improves \textbf{calibration of the treated head} relative to the production baseline, and argue this as evidence that causal models learn more generalisable representations---because they must account for the interaction between the recommendation and user behaviour, rather than user-content affinity alone.
\end{itemize}

\section{Problem Setting}

We consider a recommendation system that, for each (user, content) pair, makes a binary decision: deliver a recommendation ($T=1$) or withhold it ($T=0$). Let $\mathbf{x} \in \mathcal{X}$ denote the feature vector for a (user, content) pair. We adopt the potential outcomes framework~\cite{Rubin1974Estimating}: let $Y(1)$ denote the stream indicator if the recommendation is shown, and $Y(0)$ the stream indicator if the recommendation is withheld. The causal estimand of interest is the Conditional Average Treatment Effect (CATE),
\begin{equation}
  \tau(\mathbf{x}) = \mathbb{E}[Y(1) - Y(0) \mid \mathbf{x}],
\end{equation}
which quantifies how much the recommendation increases the probability of a stream for a given (user, content) pair. A user with $\tau(\mathbf{x}) \approx 0$ is a sure thing — they will stream with or without the recommendation, so the impression is non-incremental.

The system runs a randomised holdback experiment in which a small fraction of eligible (user, content) pairs are withheld recommendations at random. Let $\mathcal{D}_1 = \{(\mathbf{x}_i, y_i)\}$ denote the set of \emph{treated} examples (recommendation shown, $T=1$) and $\mathcal{D}_0 = \{(\mathbf{x}_j, y_j)\}$ denote the set of \emph{holdback} examples (recommendation withheld, $T=0$). Because holdback assignment is randomised, $T \perp\!\!\!\perp (Y(0), Y(1)) \mid \mathbf{x}$, so the holdback observations provide unbiased samples from the distribution of $Y(0)$.

We define the two model outputs that will be central throughout:
\begin{align}
  \hat{p}_1(\mathbf{x}) &= \hat{\mathbb{E}}[Y(1) \mid \mathbf{x}]=  \hat{\mathbb{E}}[Y \mid T=1, \mathbf{x}], \label{eq:p1} \\
  \hat{p}_0(\mathbf{x}) &= \hat{\mathbb{E}}[Y(0) \mid \mathbf{x}] = \hat{\mathbb{E}}[Y \mid T=0, \mathbf{x}]. \label{eq:p0}
\end{align}
In standard causal estimation settings, one estimates CATE as $\hat{\tau}(\mathbf{x}) = \hat{p}_1(\mathbf{x}) - \hat{p}_0(\mathbf{x})$~\cite{Shalit2017TARNet,Shi2019DragonNet,Curth2021Nonparametric}. As we discuss in Section~\ref{sec:mismatch}, this is not valid here due to a structural asymmetry in how $Y(1)$ and $Y(0)$ are defined in the production setting.

\section{The Attribution Mismatch}
\label{sec:mismatch}

The obstacle to na\"ive CATE estimation in this setting is a structural asymmetry in attribution: the outcome $Y(1)$ for treated users and the outcome $Y(0)$ for holdback users are defined using different temporal windows, and are therefore not on the same scale.

\textbf{Treated attribution.} When a recommendation is shown and the user streams the content within a short direct-response window (on the order of minutes to hours), the stream is attributed to the recommendation. This window is narrow by design---it isolates behavioral responses that are plausibly caused by the impression.

\textbf{Holdback attribution.} When a recommendation is withheld, there is no direct impression to respond to. Organic discovery is a slower process; the user might encounter the content through other surfaces over the following days. To capture this, holdback streams are attributed if they occur within a two-day window from the moment the recommendation \emph{would have} been shown. This two-day window is necessary to produce a meaningful estimate of organic behavior, but it differs from the treated attribution window.

\textbf{Why subtraction fails.} In standard causal effect estimation~\cite{Shi2019DragonNet}, $\hat{p}_1$ and $\hat{p}_0$ are estimated from treated and control observations, and CATE is recovered as their difference. This is valid only when both quantities measure the same outcome. Here they do not: the treated head is calibrated against streams within a narrow direct-response window, while the holdback head measures streams within a two-day organic window. Their difference is not an estimate of $\tau(\mathbf{x})$; it is the difference of two probabilities with different outcome definitions, and carries no clean causal interpretation. 

\textbf{Distributional mismatch} An additional practical consideration is the distributional difference between $\mathcal{D}_1$ and $\mathcal{D}_0$. In production, there is approximately a 30\% discrepancy between backend recommendation servings (the events that generate holdback labels) and client-side impressions (the events that generate treated labels), owing to latency and client-side rendering differences. The holdback and treated sets are therefore not drawn from identical distributions over $\mathbf{x}$, which is an additional reason to avoid treating $\hat{p}_0(\mathbf{x})$ as a direct counterfactual for $\hat{p}_1(\mathbf{x})$.


\textbf{What the scores can do.} Although $\hat{p}_1(\mathbf{x}) - \hat{p}_0(\mathbf{x})$ is invalid as a CATE estimator, each score individually carries well-defined decision-relevant signal. $\hat{p}_1(\mathbf{x})$ high means the recommendation is likely to drive a stream. $\hat{p}_0(\mathbf{x})$ high means the user will stream the content organically, regardless of the recommendation. The dual-threshold policy in Section~\ref{sec:method} exploits this structure directly: it uses both scores as independent decision inputs.

\section{Causal Recommendation Model}\label{sec:method}

\begin{figure}[t]
  \includegraphics[width=\columnwidth]{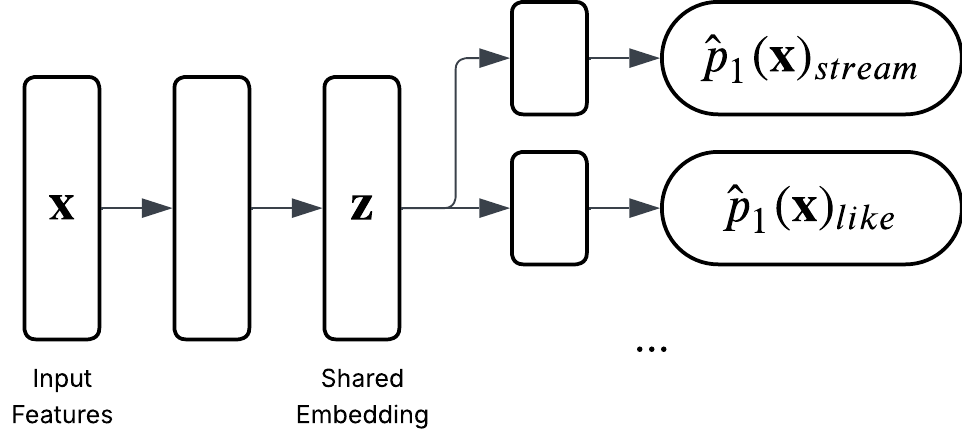}
  \caption{The baseline production recommendation model: a shared-trunk neural network with task-specific heads, trained exclusively on examples where a recommendation was shown. The primary head predicts $\hat{p}_1(\mathbf{x})$.}
  \Description{Diagram showing the baseline multi-task recommendation model architecture: a shared trunk neural network feeding into multiple task-specific output heads for stream prediction, click prediction, and other engagement tasks.}
  \label{fig:baseline}
\end{figure}

The production recommendation model is a multi-task shared-trunk neural network~\cite{Caruana1997MTL}, illustrated in Figure~\ref{fig:baseline}. A deep shared trunk maps the (user, content) feature vector $\mathbf{x}$ to a shared representation, from which task-specific heads predict various engagement outcomes (streams, clicks, etc.). The primary head, responsible for targeting decisions, is trained to predict $\hat{p}_1(\mathbf{x})$, the probability that the user streams the content after a recommendation is shown. This model is trained exclusively on treated examples $\mathcal{D}_1$, and therefore has no access to counterfactual information.

\subsection{Deep Twin Network Extension}
\label{sec:dtn}

\begin{figure}[t]
  \includegraphics[width=\columnwidth]{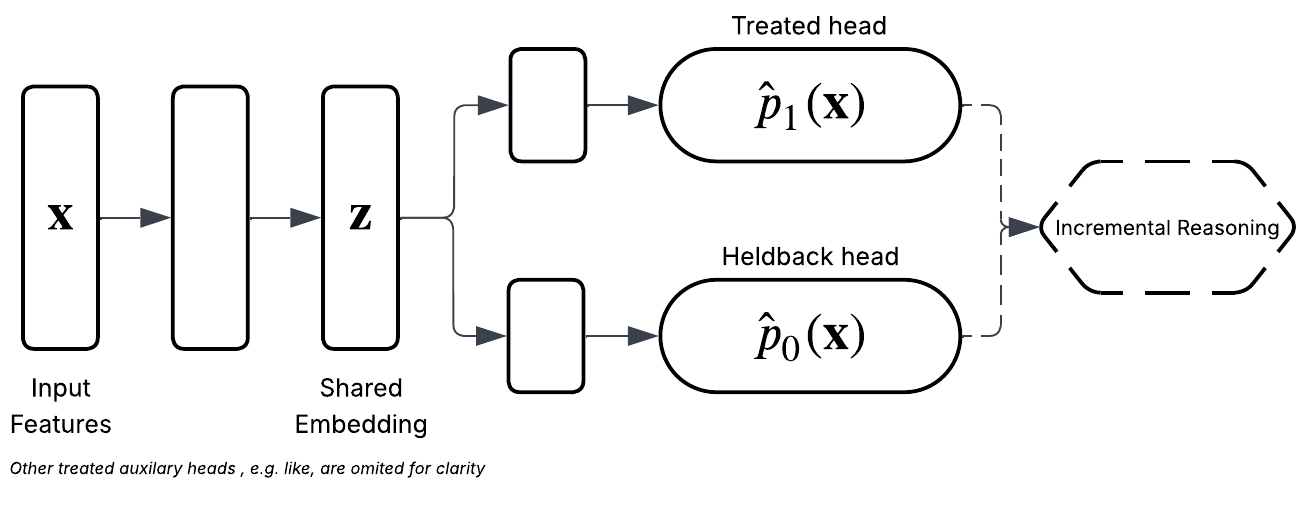}
  \caption{The causal recommendation model: the baseline extended with a holdback head. Treated examples update only the treated head and the shared trunk; holdback examples update only the holdback head and the shared trunk. The shared trunk benefits from both treatment conditions, learning representations that capture the interaction between the recommendation and user behaviour.}
  \Description{Diagram of the Deep Twin Network causal recommendation model showing the shared trunk with two output heads: a treated head for recommendation-shown predictions and a holdback head for counterfactual organic-stream predictions. Blue arrows indicate treated data flow and orange arrows indicate holdback data flow.}
  \label{fig:causal}
\end{figure}

To capture counterfactual information, we extend the baseline model to a Deep Twin Network~\cite{Vlontzos2023TwinNets} architecture by adding a holdback head that learns to predict $\hat{p}_0(\mathbf{x})$, illustrated in Figure~\ref{fig:causal}. The shared trunk is retained; the holdback head is a new output branch trained on holdback observations $\mathcal{D}_0$. This is structurally equivalent to a DragonNet~\cite{Shi2019DragonNet} with two outcome heads---one per treatment arm---but without a propensity head, since holdback assignment is randomised and the propensity is known.

Training uses a \emph{partitioned loss} in which treated examples contribute only to the treated head loss, and holdback examples contribute only to the holdback head loss, but both sets of gradients flow through the shared trunk:
\begin{equation}
  \mathcal{L} = \frac{1}{|\mathcal{D}_1|} \sum_{i \in \mathcal{D}_1} \ell_{\mathrm{bce}}(y_i, \hat{p}_1(\mathbf{x}_i)) + \frac{1}{|\mathcal{D}_0|} \sum_{j \in \mathcal{D}_0} \ell_{\mathrm{bce}}(y_j, \hat{p}_0(\mathbf{x}_j)),
  \label{eq:loss}
\end{equation}
where $\ell_{\mathrm{bce}}$ is the binary cross-entropy loss. Because both terms back-propagate through the shared trunk, the shared representation is shaped jointly by treated and holdback supervision, while each head is updated only by its own relevant examples. This partitioned structure is important: allowing holdback gradients to update the treated head (or vice versa) would conflate the two outcome definitions, which, as established in Section~\ref{sec:mismatch}, correspond to different attribution windows and cannot be directly compared.

\subsection{Dual-Threshold Targeting Policy}
\label{sec:policy}

Because $\hat{p}_1(\mathbf{x}) - \hat{p}_0(\mathbf{x})$ is not a valid CATE estimator in this setting, we instead exploit the two scores as independent binary decision inputs. We serve a recommendation only if both conditions hold:
\begin{equation}
  \pi(\mathbf{x}) = \mathbf{1}\!\left[\hat{p}_1(\mathbf{x}) \geq \theta_1 \right] \cdot \mathbf{1}\!\left[\hat{p}_0(\mathbf{x}) \leq \theta_0 \right],
  \label{eq:policy}
\end{equation}
where $\theta_1, \theta_0 \in [0,1]$ are tunable threshold parameters. The first condition ensures that the recommendation is likely to drive a stream. The second condition removes ``always-takers''---users for whom organic discovery is likely, rendering the recommendation non-incremental. Together, the two conditions identify users for whom the recommendation is both \emph{effective} (likely to produce a stream) and \emph{necessary} (the user would not stream otherwise).

The threshold $\theta_0$ provides a direct, interpretable lever on the efficiency--reach trade-off: increasing $\theta_0$ withholds more impressions from users with higher organic stream probability, reducing impression volume at the cost of potentially missing some incremental conversions; decreasing $\theta_0$ toward zero approaches the baseline policy of targeting by $\hat{p}_1(\mathbf{x})$ alone. Thresholds are set offline against randomized data to meet impression reduction targets while satisfying non-inferiority constraints on consumption metrics.

\section{Related Work}
\label{sec:related}

\textbf{Uplift modelling and CATE estimation.} The goal of identifying incremental users is closely related to uplift modelling~\cite{Radcliffe1999Uplift,Gutierrez2017Uplift,Devriendt2018Survey}, which seeks to estimate the individual-level treatment effect. Neural approaches to CATE estimation include TARNet and CFR~\cite{Shalit2017TARNet}, DragonNet~\cite{Shi2019DragonNet}, Deep Twin Networks~\cite{Vlontzos2023TwinNets}, and a range of meta-learning approaches~\cite{Curth2021Nonparametric,Curth2021Inductive}. The present work is distinguished by its production setting: we operate on a live system with tens of millions of users, with the additional structural complexity that treated and holdback outcomes are not measured on the same scale — a challenge not addressed in the benchmark-dataset literature for these methods. Recent theoretical work has further studied the fundamental difficulty of validating causal models against experimental data~\cite{Fawkes2025Hardness}, underscoring the importance of careful experimental design in production deployments such as ours.

\textbf{Causal recommendation systems.} The use of causal reasoning to improve recommendation has been studied through inverse propensity score weighting for debiasing~\cite{Schnabel2016Debiasing}, counterfactual offline evaluation~\cite{Gilotte2018Offline}, and treatment-aware modelling~\cite{Liang2016Causal}. Contrastive approaches to learning causally relevant treatment representations have also been explored~\cite{Corcoll2026Contrastive}. The dominant thread in this literature targets debiasing click and conversion prediction; our work differs in that we explicitly target incrementality rather than correcting for selection bias in the observation of engagement.

\textbf{Impression efficiency.} The problem of spending impressions on always-takers is well known in direct marketing~\cite{Radcliffe1999Uplift,Devriendt2018Survey} and has begun to attract attention in recommendation~\cite{Ma2018Entire,Chapelle2014Simple}. The contribution of this work is to demonstrate that a production-scale incremental recommendation system can be built from existing holdback infrastructure, and that doing so achieves meaningful impression savings without degrading the recommendation impact.

\section{Experiments}
\label{sec:experiments}


We evaluate the dual-threshold policy in a live A/B test on Spotify's Home recommendation surface (Figure~\ref{fig:ui}). The test was run at production scale with millions of users. Operating at this scale required the causal model to match the latency and throughput constraints of the production serving infrastructure.

The experiment comprised three arms:

\begin{itemize}
  \item \textbf{Control}: The production recommendation model, targeting by treated-stream score $\hat{p}_1(\mathbf{x})$ trained on $\mathcal{D}_1$ only.
  \item \textbf{Treatment-Model}: The causal model of Section~\ref{sec:dtn}, targeting by $\hat{p}_1(\mathbf{x})$ using the filtering for just that arm: $\mathbf{1}\!\left[\hat{p}_1(\mathbf{x}) \geq \theta_1\right]$. This arm isolates the effect from training the shared trunk on $\mathcal{D}_1$ and $\mathcal{D}_0$.
  \item \textbf{Treatment-Causal}: The causal model with the dual-threshold policy $\pi(x)$ of Section~\ref{sec:policy}. 
\end{itemize}

The primary comparison of interest is Treatment-Model versus Treatment-Causal, which isolates the effect of the dual-threshold policy while holding the model architecture and training data fixed.


We track two metrics over two weeks in the test. \textbf{Recommendation impressions per user}: the primary efficiency metric. \textbf{Consumption minutes of recommended content}: the primary impact metric, with non-inferiority required. 


Table~\ref{tab:results} summarises the outcomes of the Treatment-Model versus Treatment-Causal comparison.

\begin{table}[h]
\caption{A/B test: Treatment-Causal vs.\ Treatment-Model.} 
\label{tab:results}
\small
\begin{tabular}{p{4.5cm}p{1cm}p{2.0cm}}
\toprule
\textbf{Metric} & \textbf{$\Delta$} & \textbf{$95\%$ CI} \\
\midrule
Recommendation impressions & $-7.1\%$ & $[-7.2\%,-6.9\%]$ \\
Recommended content consumption & $-0.37\%$ & $[-0.86\%,+0.21\%]$ \\
\bottomrule
\end{tabular}
\end{table}

The dual-threshold policy reduces recommendation impressions by 7\% relative to targeting with the treated-stream score alone, while producing no statistically significant change in the consumption of recommended content or in listener satisfaction guardrails. 

The magnitude of the impression reduction is informative. It implies that approximately 93\% of the impressions delivered by the production recommendation model are already incremental: users who would not have discovered the content without the recommendation. This is a consequence of the organic discovery landscape on the platform---for the content types featured on the Home surface, the recommendation itself is one of very few reliable discovery mechanisms available to users who do not already closely follow the relevant artists or podcasters. The dual-threshold policy successfully identifies and removes the non-incremental 7\%, demonstrating precise targeting and removal of always-takers.

\section{Calibration and Generalisation}
\label{sec:calibration}

A benefit of the causal architecture is improved calibration of the treated head $\hat{p}_1(\mathbf{x})$. Calibration matters for threshold-based policies: if the model's predicted probabilities do not accurately reflect empirical stream frequencies, the threshold parameters $\theta_1$ and $\theta_0$ lose their interpretation and the policy cannot be reliably tuned.

\begin{figure}[t]
  \begin{minipage}[t]{0.48\columnwidth}
    \includegraphics[width=\linewidth]{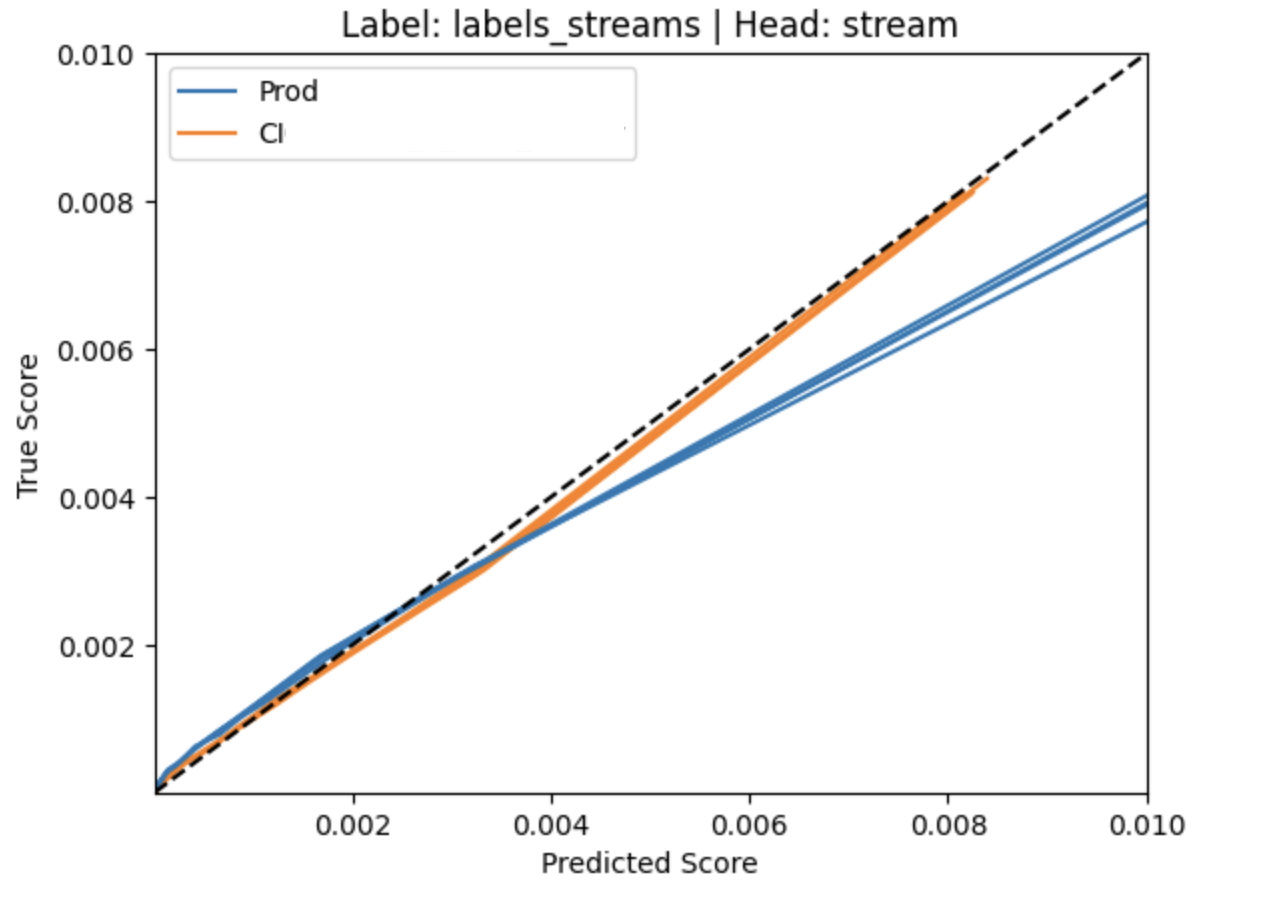}
  \end{minipage}\hfill
  \begin{minipage}[t]{0.48\columnwidth}
    \includegraphics[width=\linewidth]{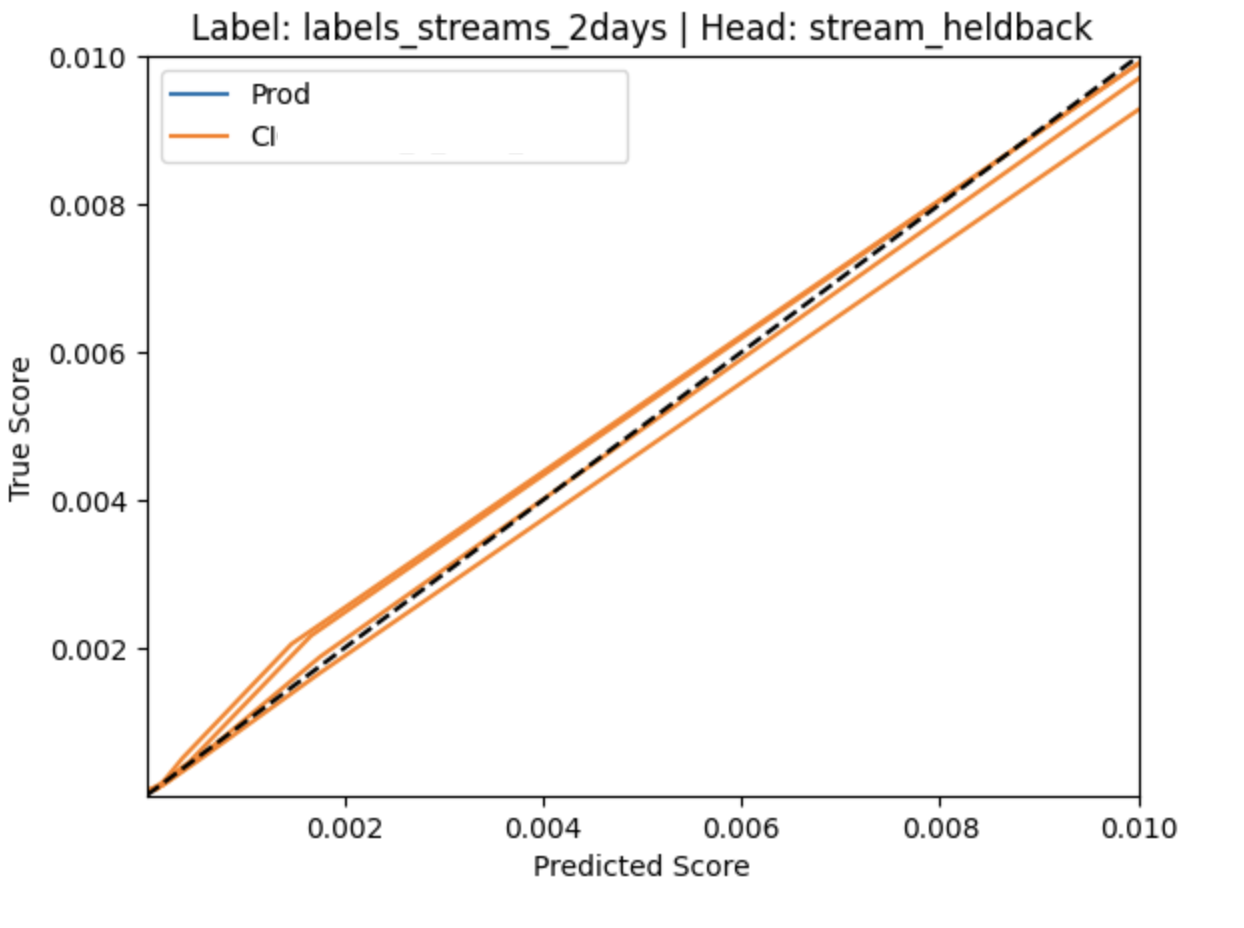}
  \end{minipage}
  \caption{Calibration curves for (left) the treated head and (right) the holdback head. Left: the causal treated head (orange) tracks the diagonal more closely than the production baseline (blue), with the largest improvement at higher predicted probabilities. Right: the holdback head is well-calibrated, with minor deviations at the tail attributable to the smaller holdback sample size.}
  \Description{Two calibration plots side by side. Left plot shows mean predicted probability vs observed frequency for the causal treated head (orange line) and production baseline (blue line), with the causal head tracking closer to the diagonal. Right plot shows the holdback head calibration curve near the diagonal.}
  \label{fig:calibration}
\end{figure}

Figure~\ref{fig:calibration} shows calibration curves for the treated head and the holdback head. The causal treated head (orange) tracks the diagonal more closely than the production baseline (blue), with the most pronounced improvement at higher predicted probabilities.

We hypothesise that this calibration improvement reflects a genuine generalisation benefit of the causal architecture. A model trained exclusively on treated data learns a mapping from user-content features to stream probability \emph{conditional on a recommendation being shown}. This conflates two distinct factors: the user's affinity for the content, and the effect of the recommendation itself. A model that cannot disentangle these factors will tend to over-predict for high-affinity users (who would stream regardless). 

Joint training with holdback data, through the shared trunk, forces the model to represent both \emph{affinity} and \emph{recommendation sensitivity}---the degree to which the recommendation changes the user's behavior. The holdback head must predict organic stream probability accurately, and the shared trunk must therefore encode the features that predict behaviour in the absence of a recommendation. This richer representation generalises better across treatment conditions: the treated head, sharing the trunk, inherits a more complete picture of user-content affinity.

This observation connects to a broader principle: causal models, by virtue of being trained on data from multiple treatment conditions, learn representations that are more robust to distribution shift than models trained on a single observational regime~\cite{Fawkes2025Hardness,Corcoll2026Contrastive}. The calibration improvement here is an empirical instance of this generalisation benefit in a production recommendation setting.

The holdback head (Figure~\ref{fig:calibration}, right) is also well-calibrated, with minor deviations at higher predicted probabilities attributable to the smaller sample size of holdback data relative to treated data.

\section{Conclusion}

We have shown how an existing production recommendation model with holdback infrastructure can be transformed into a causal incremental targeting system. 
In a production-scale A/B test on millions of Spotify users, this model reduces recommendation impressions by 7\% with no statistically significant reduction in recommended content consumption or user satisfaction---increasing incremental impact. We further demonstrate that joint training with holdback data improves model calibration, arguing this as evidence that the causal model learns more generalisable representations.


\begin{acks}
The authors thank the Kipp team at Spotify for infrastructure support and experimental design, and the Advanced Causal Inference lab at Spotify for helpful discussions.
\end{acks}

\bibliographystyle{ACM-Reference-Format}
\bibliography{refs}

\end{document}